\documentclass[letterpaper, 10 pt, journal, twoside]{IEEEtran}
\usepackage{epsfig} 
\usepackage{amsmath} 
\usepackage{amssymb}  
\usepackage{color}
\usepackage{cite}
\usepackage[linesnumbered,ruled]{algorithm2e}
\usepackage[normalem]{ulem} 
\makeatletter
\let\NAT@parse\undefined
\makeatother
\usepackage{hyperref}
\hypersetup{pdfstartview=FitH,
            colorlinks=true,
            linkcolor=red,
            citecolor=black
            }
\usepackage{booktabs}
\usepackage{arydshln}
\usepackage{multirow}
\usepackage{mathrsfs}
\usepackage[caption=false,font=footnotesize]{subfig}
\usepackage[dvipsnames]{xcolor}
\usepackage{subcaption}
\usepackage{graphicx}
\usepackage{subfig} 
\usepackage{soul}
\usepackage{ulem}
\newcounter{RNum}
\renewcommand{\theRNum}{\arabic{RNum}}
\newcommand{\Remark}{\noindent\textit{\textbf{Remark}~\refstepcounter{RNum}\textbf{\theRNum}: }}
\newcommand{\NoOne}[1]{\textcolor{red}{#1}}
\newcommand{\NoTwo}[1]{\textcolor{green}{#1}}
\newcommand{\NoThree}[1]{\textcolor{blue}{#1}}
\usepackage{xcolor}
\soulregister\NoOne7
\soulregister\NoTwo7
\soulregister\NoThree7
\soulregister\Remark7

\soulregister{\cite}7 
\soulregister{\citep}7 
\soulregister{\citet}7 
\soulregister{\ref}7
\soulregister{\pageref}7 
\usepackage{amssymb}  
\usepackage{pifont}   
\usepackage[absolute,overlay]{textpos}
\usepackage{multirow}
\usepackage{multicol}
\usepackage{graphicx}
\usepackage{amsmath}
\usepackage{amssymb}
\usepackage{booktabs}
\usepackage{xcolor,colortbl}
\usepackage{makecell}
\usepackage{pifont}
\usepackage{balance}
\usepackage{cancel}
\usepackage{adjustbox}

\definecolor{nbarrier}{RGB}{255, 120, 50}
\definecolor{nbicycle}{RGB}{255, 192, 203}
\definecolor{nbus}{RGB}{255, 255, 0}
\definecolor{ncar}{RGB}{0, 150, 245}
\definecolor{nconstruct}{RGB}{0, 255, 255}
\definecolor{nmotor}{RGB}{200, 180, 0}
\definecolor{npedestrian}{RGB}{255, 0, 0}
\definecolor{ntraffic}{RGB}{255, 240, 150}
\definecolor{ntrailer}{RGB}{135, 60, 0}
\definecolor{ntruck}{RGB}{160, 32, 240}
\definecolor{ndriveable}{RGB}{255, 0, 255}
\definecolor{nother}{RGB}{139, 137, 137}
\definecolor{nsidewalk}{RGB}{75, 0, 75}
\definecolor{nterrain}{RGB}{150, 240, 80}
\definecolor{nmanmade}{RGB}{213, 213, 213}
\definecolor{nvegetation}{RGB}{0, 175, 0}

\definecolor{car}{rgb}{0.39215686, 0.58823529, 0.96078431}
\definecolor{bicycle}{rgb}{0.39215686, 0.90196078, 0.96078431}
\definecolor{motorcycle}{rgb}{0.11764706, 0.23529412, 0.58823529}
\definecolor{truck}{rgb}{0.31372549, 0.11764706, 0.70588235}
\definecolor{other-vehicle}{rgb}{0.39215686, 0.31372549, 0.98039216}
\definecolor{person}{rgb}{1.        , 0.11764706, 0.11764706}
\definecolor{bicyclist}{rgb}{1.        , 0.15686275, 0.78431373}
\definecolor{motorcyclist}{rgb}{0.58823529, 0.11764706, 0.35294118}
\definecolor{road}{rgb}{1.        , 0.        , 1.        }
\definecolor{parking}{rgb}{1.        , 0.58823529, 1.        }
\definecolor{sidewalk}{rgb}{0.29411765, 0.        , 0.29411765}
\definecolor{other-ground}{rgb}{0.68627451, 0.        , 0.29411765}
\definecolor{building}{rgb}{1.        , 0.78431373, 0.        }
\definecolor{fence}{rgb}{1.        , 0.47058824, 0.19607843}
\definecolor{vegetation}{rgb}{0.        , 0.68627451, 0.        }
\definecolor{trunk}{rgb}{0.52941176, 0.23529412, 0.        }
\definecolor{terrain}{rgb}{0.58823529, 0.94117647, 0.31372549}
\definecolor{pole}{rgb}{1.        , 0.94117647, 0.58823529}
\definecolor{traffic-sign}{rgb}{1.        , 0.        , 0.    }    

\makeatletter
\newcommand{\car@semkitfreq}{3.92}
\newcommand{\bicycle@semkitfreq}{0.03}
\newcommand{\motorcycle@semkitfreq}{0.03}
\newcommand{\truck@semkitfreq}{0.16}
\newcommand{\othervehicle@semkitfreq}{0.20}
\newcommand{\person@semkitfreq}{0.07}
\newcommand{\bicyclist@semkitfreq}{0.07}
\newcommand{\motorcyclist@semkitfreq}{0.05}
\newcommand{\road@semkitfreq}{15.30}  %
\newcommand{\parking@semkitfreq}{1.12}
\newcommand{\sidewalk@semkitfreq}{11.13}  %
\newcommand{\otherground@semkitfreq}{0.56}
\newcommand{\building@semkitfreq}{14.1}  %
\newcommand{\fence@semkitfreq}{3.90}
\newcommand{\vegetation@semkitfreq}{39.3}  %
\newcommand{\trunk@semkitfreq}{0.51}
\newcommand{\terrain@semkitfreq}{9.17} %
\newcommand{\pole@semkitfreq}{0.29}
\newcommand{\trafficsign@semkitfreq}{0.08}
\newcommand{\semkitfreq}[1]{{\csname #1@semkitfreq\endcsname}}
\newcommand{\name}{\textsc{Omega}\xspace}

\usepackage{tikz,xcolor,hyperref}
\definecolor{lime}{HTML}{A6CE39}
\DeclareRobustCommand{\orcidicon}{%
    \begin{tikzpicture}
    \draw[lime, fill=lime] (0,0) 
    circle [radius=0.16] 
    node[white] {
   {\fontfamily{qag}\selectfont \tiny ID}};    \draw[white, fill=white] (-0.0625,0.095) 
    circle [radius=0.007];    \end{tikzpicture}
    \hspace{-2mm}}
\foreach \x in {A, ..., Z}{%
    \expandafter\xdef\csname orcid\x\endcsname{\noexpand\href{https://orcid.org/\csname orcidauthor\x\endcsname}{\noexpand\orcidicon}}
    }

\hypersetup{colorlinks = true, 
	    linkcolor = blue,
            citecolor = blue}

\begin{document}

\title{\name: Efficient Occlusion-Aware Navigation for Air-Ground Robots in Dynamic Environments via State Space Model}

\author{Junming Wang\orcidA{}, Xiuxian Guan\orcidC{}, Zekai Sun\orcidB{}, Tianxiang Shen\orcidD{}, Dong Huang, Fangming Liu\orcidE{},~\IEEEmembership{Senior Member,~IEEE} and Heming Cui\orcidF{},~\IEEEmembership{Member,~IEEE}
\thanks{Received: July 3, 2024; Revised: October 3, 2024; Accepted: December 3, 2024. This paper was recommended for publication by Associate Editor Asfour, Tamim and
Vasseur, Pascal upon evaluation of the reviewers’ comments. This work is supported in part by National Key R\&D Program of China (2022ZD0160201), HK RGC RIF (R7030-22), HK ITF (GHP/169/20SZ), a Huawei Flagship Research Grant in 2023, HK RGC GRF (Ref: 17208223 \& 17204424), and the HKU-CAS Joint Laboratory for Intelligent System Software. \textit{(Corresponding author: Heming Cui.)} 

Junming Wang, Xiuxian Guan, Zekai Sun, Tianxiang Shen, Dong Huang are with the University of Hong Kong, Hong Kong SAR 999077, China. (e-mail: jmwang@cs.hku.hk). 

Fangming Liu is with Peng Cheng Laboratory, and Huazhong University of Science and Technology, Wuhan 430074, China. (e-mail: fmliu@hust.edu.cn). 

Heming Cui is with the University of Hong Kong, Hong Kong SAR 999077, China, and also with the the Shanghai AI Labolatory, , Shanghai 200232, China. (e-mail: heming@cs.hku.hk). 

Code and Video Page: \url{https://jmwang0117.github.io/OMEGA/}.
}

\thanks{Digital Object Identifier (DOI): see top of this page.}}

\markboth{IEEE ROBOTICS AND AUTOMATION LETTERS. PREPRINT VERSION. ACCEPTED DECEMBER, 2024}
{Wang \MakeLowercase{\textit{et al.}}: OMEGA: Efficient Occlusion-Aware Navigation for Air-Ground Robots in Dynamic Environments via State Space Model}

\maketitle

\begin{abstract}

Air-ground robots (AGRs) are widely used in surveillance and disaster response due to their exceptional mobility and versatility (i.e., flying and driving). Current AGR navigation systems perform well in static occlusion-prone environments (e.g., indoors) by using 3D semantic occupancy networks to predict occlusions for complete local mapping and then computing Euclidean Signed Distance Field (ESDF) for path planning. However, these systems face challenges in dynamic scenes (e.g., crowds) due to limitations in perception networks' low prediction accuracy and path planners' high computation overhead. In this paper, we propose \name, which contains \underline{O}cc\underline{M}amba with an \underline{E}fficient A\underline{G}R-pl\underline{A}nner to address the above-mentioned problems. OccMamba adopts a novel architecture that separates semantic and occupancy prediction into independent branches, incorporating the mamba module to efficiently extract semantic and geometric features in 3D environments. This ensures the network can learn long-distance dependencies and improve prediction accuracy. Features are then combined within the Bird's Eye View (BEV) space to minimise computational overhead during feature fusion. The resulting semantic occupancy map is integrated into the local map, providing occlusion awareness of the dynamic environment. Our AGR-Planner utilizes this local map and employs Kinodynamic A* search and gradient-based trajectory optimization for ESDF-free and energy-efficient planning. Experiments demonstrate that OccMamba outperforms the state-of-the-art 3D semantic occupancy network with 25.0\% mIoU. End-to-end navigation experiments in dynamic scenes verify OMEGA's efficiency, achieving a 96\% average planning success rate. 

\end{abstract}


%
\IEEEpeerreviewmaketitle

\begin{figure}[t]
  \centering
     \includegraphics[width=0.9\linewidth]{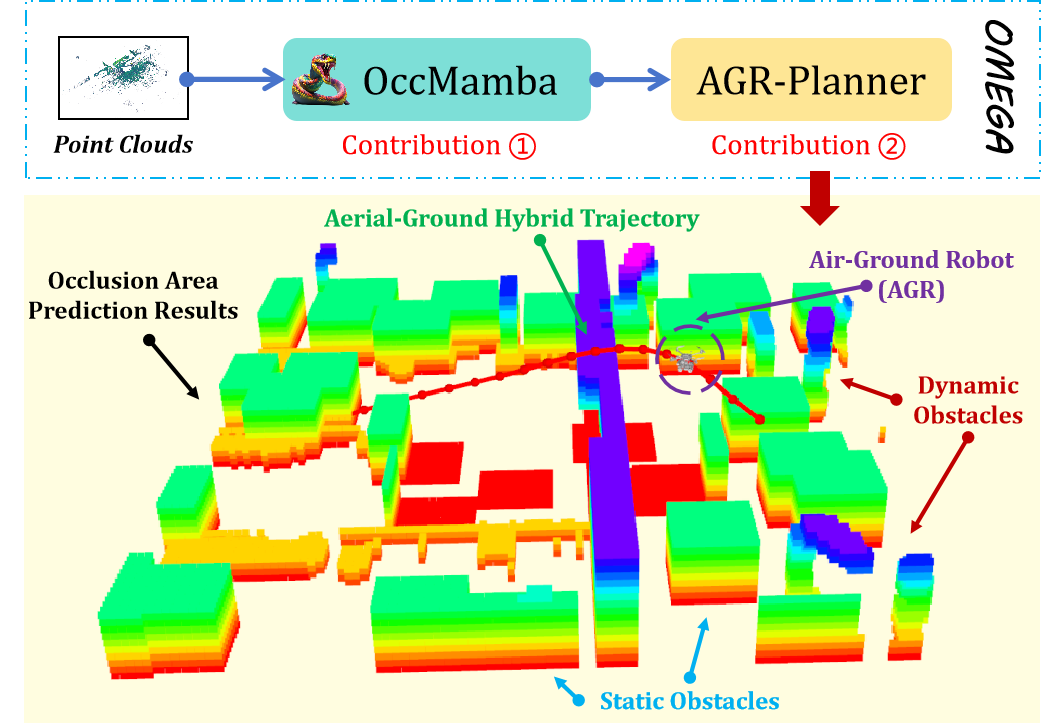}
   \caption{\small  \name is the first AGR-specific navigation system that enables occlusion-aware mapping and pathfinding in dynamic scenarios. It integrates OccMamba for real-time obstacle prediction from point clouds and updates local maps accordingly, while AGR-Planner rapidly generates reliable paths using the updated local map.}
   \label{fig:head}
\end{figure}

\section{INTRODUCTION}

\IEEEPARstart{I}{n} recent years, air-ground robots (AGRs) have attracted significant attention from both academia \cite{fan2019autonomous, zhang2022autonomous,zhang2023model, wang2024agrnav} and industry due to their versatile navigation capabilities in the ground and aerial domains. To enhance AGR navigation in occlusion-prone environments, existing works employ sensors, such as depth cameras, to collect point clouds. These point clouds are then processed by 3D semantic occupancy networks \cite{cao2022monoscene,cheng2021s3cnet,tang2024sparseocc,jiang2023symphonize}, which predict occluded areas and generate a complete local map. Based on the local map, an Euclidean Signed Distance Field (ESDF) map \cite{zhang2022autonomous, wang2024agrnav} is constructed for hybrid aerial-ground path planning. 

Unfortunately, while these methods have shown success in static occlusion-prone environments (e.g., indoor or forest), they struggle in dynamic environments with moving obstacles (Fig.~\ref{fig:head}). This limitation can be attributed to two main factors: the \textit{perception network's} high inference latency and low prediction accuracy, and the \textit{path planner's} high computational overhead. Firstly, despite the improved accuracy of transformer-based 3D semantic occupancy prediction networks \cite{zhang2023occformer,li2023voxformer} in simultaneously predicting static and dynamic obstacles for occlusion-free local mapping, their high computational overhead hinders deployment on resource-constrained AGR devices (e.g., Jetson Xavier NX). Secondly, current AGR path planners \cite{zhang2022autonomous, wang2024agrnav} allocate nearly 70\% of the total local planning time to construct the Euclidean Signed Distance Field (ESDF) map. This extensive processing time is impractical for dynamic environments with rapidly changing scenes, as it heightens the risk of collisions (Table~\ref{tab:head}).

Our key insight to address the above limitations is to propose an efficient 3D semantic occupancy network that ensures real-time inference while improving prediction accuracy. We draw inspiration from state space models (SSMs) \cite{gu2021efficiently} and their improved versions, such as Mamba \cite{dao2024transformers,gu2023mamba}. By integrating mamba blocks into a new 3D semantic occupancy network architecture, we aim to enable the network to model long-distance dependencies and perform parallel feature learning. This approach facilitates real-time occlusion prediction in highly dynamic environments, resulting in more complete local maps. Furthermore, while \textit{Zhou et al.} \cite{zhou2021ego} developed an ESDF-free path planner for quadcopters, it does not address AGR-specific requirements in dynamic scenarios, particularly energy efficiency and dynamic constraints. Consequently, exploring ESDF-free AGR path planners is crucial to improving overall navigation performance.

\begin{table}[t]
\centering
\caption{Comparison with baseline AGR systems.}
\resizebox{\columnwidth}{!}{%
\begin{tabular}{@{}lccccc@{}}
\toprule
Method & Dyn. Env. & Occl. Aware & Mov. Time & Succ. Rate & Energy \\ \midrule
HDF \cite{fan2019autonomous} & \ding{55} & \ding{55} & \ding{55} & \ding{55} & \ding{55} \\
TABV \cite{zhang2022autonomous} & \ding{55} & \ding{55} & \ding{55} & \ding{55} & \ding{55} \\
M-TABV \cite{zhang2023model} & \ding{55} & \ding{55} & \ding{55} & \ding{55} & \ding{55} \\
AGRNav \cite{wang2024agrnav} & \ding{55} & \checkmark & \ding{55} & \ding{55} & \checkmark \\
HE-Nav \cite{wang2024he} & \ding{55} & \checkmark & \checkmark & \ding{55} & \checkmark \\
\midrule 
\textbf{\name (Ours)} & \checkmark & \checkmark & \checkmark & \checkmark & \checkmark \\
\bottomrule
\end{tabular}%
}
\label{tab:head}
\end{table}

\begin{figure*}[t]
  \centering
     \includegraphics[width=0.95\linewidth]{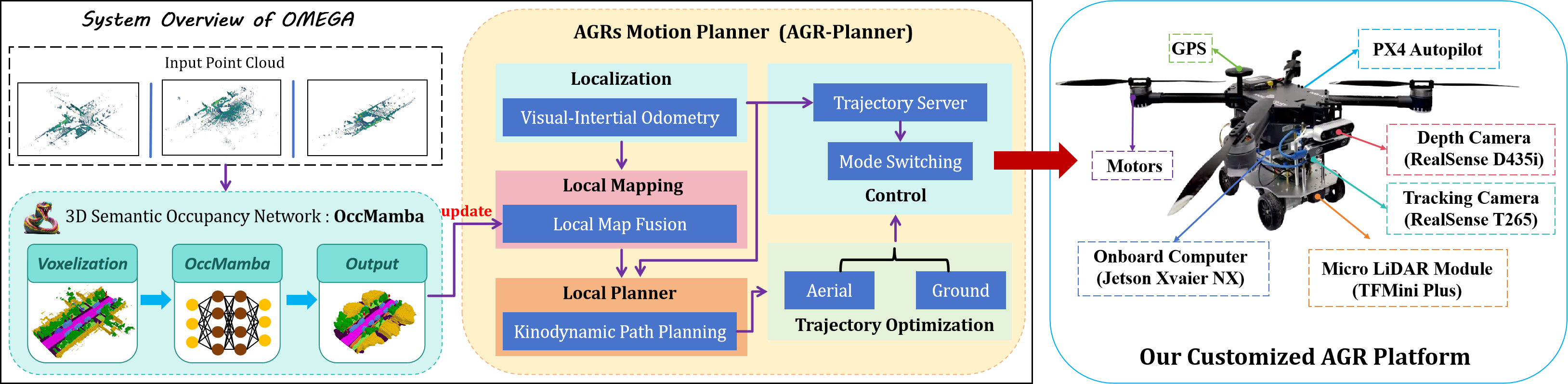}
   \caption{\small \name system architecture. The perception network (i.e., OccMamba) and AGR-planner run asynchronously on the onboard computer, connected through a query-based map update method from \cite{wang2024agrnav} to ensure real-time local map updates with prediction results.  }
   \label{fig:overview}
\end{figure*}

Based on these insights, We introduce \name (Fig.~\ref{fig:overview}), consisting of two key components: OccMamba and AGR-planner. OccMamba, the first 3D semantic occupancy network based on state-space models (SSMs), departs from previous networks that jointly learn semantics and occupancy by separating these predictions into different branches (Fig.~\ref{fig:occmamba}). This separation enables specialized learning within each domain, improving prediction accuracy and fully leveraging the complementary properties of semantic and geometric features in the subsequent feature fusion stage. We also integrate novel Sem-Mamba and Geo-Mamba blocks into these branches to capture long-distance dependencies critical for semantic accuracy and occupancy prediction. By projecting features into the bird's-eye view (BEV) space, we reduce fusion latency to achieve real-time inference. Prediction results are merged into the local map using a low-latency update method from \cite{wang2024agrnav}, ensuring an up-to-date and accurate map in highly dynamic environments.

During the planning phase, we propose AGR-Planner, which builds on EGO-Swarm \cite{zhou2021ego} for aerial-ground hybrid path planning. To address AGR-specific requirements, particularly energy efficiency and dynamic constraints, we add additional energy costs to motion primitives involving aerial destinations, promoting energy efficiency. Simultaneously, we account for AGRs' non-holonomic constraints by limiting ground control point curvature. Finally, by integrating the obstacle distance estimation method from \cite{zhou2021ego} and backend trajectory optimization, AGR-Planner generates ESDF-free, energy-efficient, smooth, collision-free, and dynamically feasible trajectories. 

We first assessed OccMamba on the SemanticKITTI benchmark, comparing its accuracy and inference speed to some leading occupancy networks. Then, we tested \name (Fig.~\ref{fig:overview}) in simulated and real dynamic environments, contrasting it with two solid and open-source AGR navigation baselines, showcasing its superior efficiency. Our evaluation reveals:

\begin{itemize}
    \item \textbf{OccMamba is efficient and real-time.} OccMamba achieves state-of-the-art performance (mIoU = 25.0) on the SemanticKITTI benchmark and enables high-speed inference (i.e., 22.1 FPS). (§~\ref{sec:B1})
    
    \item \textbf{\name is efficient.} \name achieved success rates of 98\% in the simulation scenarios while having the shortest average movement time (i.e., 16.1s).  (§~\ref{sec:C1})
    
    \item \textbf{\name is energy-saving.} The results of real dynamic environment navigation show that \name can save about 18\% of energy consumption. (§~\ref{sec:D1})
    
\end{itemize}

\section{RELATED WORK}

\subsection{Dynamic Navigation System for AGRs}
Researchers have explored various aerial-ground robot configurations, such as incorporating passive wheels \cite{zhang2022autonomous,wu2023unified,pan2023skywalker}, cylindrical cages \cite{kalantari2013design}, or multi-limb \cite{martynov2023morphogear} onto drones. Recently, \textit{Fan et al.} \cite{fan2019autonomous} address ground-aerial motion planning. Their approach initially employs the A* algorithm to search for a geometric path as guidance, favouring ground paths by adding extra energy costs to aerial paths. \textit{Zhang et al.} \cite{zhang2022autonomous} proposed a path planner and controller capable of path searching, but it relies on an ESDF map. The intensive computation and limited perception of occluded areas lead to a low success rate in path planning and increased energy consumption. \textit{Wang et al.} \cite{wang2024agrnav} proposed AGRNav, the first AGR navigation system with occlusion perception capability. Although it performs well in static environments, its simple perception network structure design and the defects of the path planner make it difficult to operate in complex and changing dynamic environments. Our work aims to explore new AGR navigation systems for efficient navigation in dynamic environments.

\subsection{3D Semantic Occupancy Prediction}
3D semantic occupancy prediction is crucial for interpreting occluded environments, as it discerns the spatial layout beyond visual obstructions by merging geometry with semantic clues. This process enables autonomous systems to anticipate hidden areas, crucial for safe navigation and decision-making. Research on 3D semantic occupancy prediction can be summarized into three main streams: \textit{Camera-based} approaches capitalize on visual data, with pioneering works like MonoScene by Cao et al.~\cite{cao2022monoscene} exploiting RGB inputs to infer indoor and outdoor occupancy. Another notable work by \textit{Li et al.} \cite{li2023voxformer} is VoxFormer, a transformer-based semantic occupancy framework capable of generating complete 3D volume semantics using only 2D images. \textit{LiDAR-based} approaches like S3CNet by Cheng et al.~\cite{cheng2021s3cnet}, JS3C-Net by Yan et al.~\cite{yan2021sparse}, and SSA-SC by Yang et al.~\cite{yang2021semantic}, which adeptly handle the vastness and variability of outdoor scenes via point clouds. \textit{Fusion-based} approaches aim to amalgamate the contextual richness of camera imagery with the spatial accuracy of LiDAR data. The Openoccupancy benchmark by Wang et al.~\cite{wang2023openoccupancy} is a testament to this synergy, providing a platform to assess the performance of integrated sensor approaches. 

\subsection{State Space Models (SSMs) and Mamba}

State-Space Models (SSMs), including the S4 model \cite{gu2021efficiently}, have proven effective in sequence modelling, particularly for capturing long-range dependencies, outperforming traditional CNNs and Transformers. The Mamba model \cite{gu2023mamba,dao2024transformers} builds on this by introducing data-dependent SSM layers, offering enhanced adaptability and significantly improved efficiency in processing long sequences. This has led Mamba to excel in various domains, including robot manipulation \cite{liu2024robomamba}, point cloud analysis \cite{liang2024pointmamba,zhu2024vision}, and video understanding \cite{li2024videomamba}, due to its scalability and versatility. Leveraging Mamba's strengths, we developed OccMamba, the first mamba-based 3D semantic occupancy prediction network to learn long-distance dependencies for understanding dynamic scenes.

\begin{figure*}[t]
  \centering
     \includegraphics[width=0.95\linewidth]{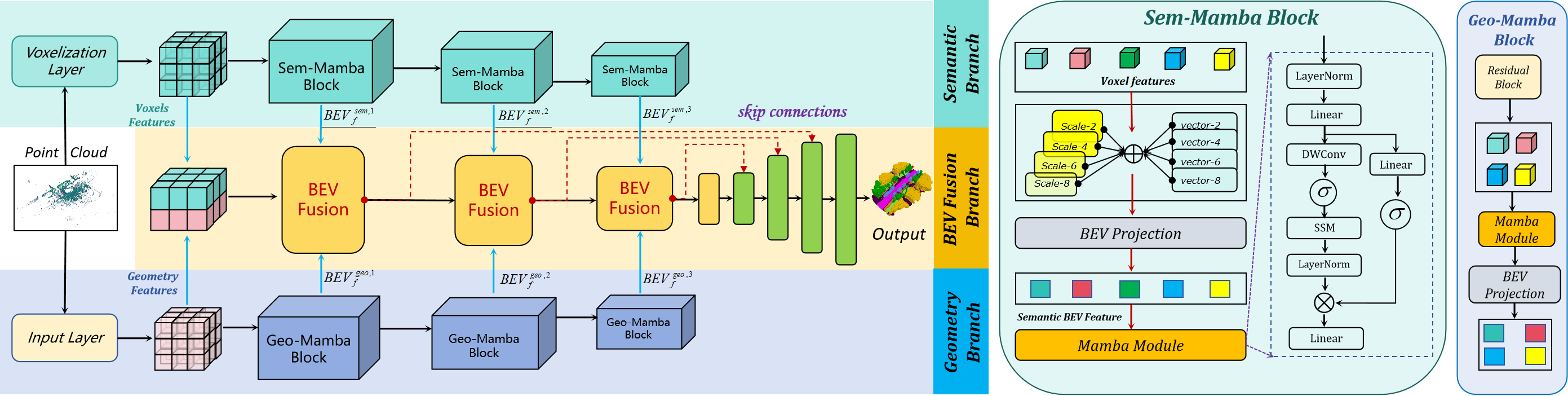}
   \caption{\small The overview of the proposed OccMamba. It consists of semantic, geometry and BEV fusion branches. Meanwhile, lightweight MLPs serve as auxiliary heads during training, attached after each encoder block in the semantic and completion branches for voxel predictions. At the inference stage, these heads are detached to preserve a streamlined network architecture. }
   \label{fig:occmamba}
\end{figure*}

\section{3D Semantic Occupancy Network of \name}
\label{sec:3}
OccMamba (Fig.~\ref{fig:occmamba}) features three branches: semantic, geometric, and BEV fusion. The semantic and geometric branches are supervised by multi-level auxiliary losses, which are removed during inference. Multi-scale features generated by these two branches are fused in the BEV space to alleviate the computational overhead caused by dense feature fusion.

\subsection{OccMamba Network Structure}
\noindent\textbf{\textit{Semantic Branch}:} The semantic branch consists of a voxelization layer and three encoder Sem-Mamba Blocks with the same structures. The input point cloud $P \in \mathbb{R}^{N \times 3}$ is converted to a multi-scale voxel representation at a voxel resolution $s$ by the voxelization layer. These voxels are then aggregated by maximum pooling to obtain a unified feature vector for each voxel. The vectors from different scales are merged to form the final voxel feature ${V}_{f_m}$, with a size of $L \times W \times H$, where $f_m$ represents the voxel index. The semantic features $\{S_{f}^{1},S_{f}^{2},S_{f}^{3} \}$ are projected into the bird's-eye view (BEV) space by assigning a unique BEV index to each voxel based on its $f_m$ value. Features sharing the same BEV index are aggregated using max pooling, resulting in sparse BEV features. These sparse features are then densified using the densification function of Spconv \cite{spconv2022} to produce dense BEV features $\{BEV_{f}^{sem,1}, BEV_{f}^{sem,2}, BEV_{f}^{sem,3}\}$. 

\noindent\textbf{\textit{Sem-Mamba Block}:} Mamba \cite{gu2023mamba,dao2024transformers}, a selective state-space model, has recently outperformed CNN- and Transformer-based approaches in various vision tasks due to its efficient long-distance sequence modelling and linear-time complexity. These characteristics make Mamba promising for improving 3D semantic occupancy prediction accuracy while maintaining fast reasoning in dynamic environments. Inspired by mamba's success, we introduce the Sem-Mamba block as the semantic branch encoder in our network to enable efficient semantic representation learning. Specifically, state space models \cite{gu2021efficiently} introduce hidden states $h(t) \in \mathbb{R} ^N$ to map inputs $x(t) \in \mathbb{R} ^L$ to outputs $y(t)$, with the continuous state space dynamics governed by:

\begin{equation}
h^{'}(t) =\mathbf{A} h(t)+\mathbf{B}x(t), y(t)=\mathbf{C}h(t)
\end{equation}
Using a time scale parameter $\bigtriangleup$, the Mamba model discretizes the continuous parameters, yielding the discretized state space equations:
\begin{equation}
\overline{\mathbf{A} } =exp(\bigtriangleup \mathbf{A} ), \overline{\mathbf{B} } =(\bigtriangleup\mathbf{A} )^{-1}(exp\bigtriangleup( \mathbf{A})-\mathbf{I} )\cdot \bigtriangleup \mathbf{B}
\end{equation}
\begin{equation}
h(t)=\overline{\mathbf{A} } h_{t-1}+ \overline{\mathbf{B} } x_t , y_t=\mathbf{C} h_t
\end{equation}
The global convolution kernel $\overline{\mathbf{K} } \in \mathbb{R} ^L$ is used to calculate the output $y$:
\begin{equation}
\overline{\mathbf{K} } =(C\overline{B} ,C\overline{AB},...,C\overline{A}^{m-1}\overline{B} ), y=x * \overline{\mathbf{K} }
\end{equation}

In each Sem-Mamba Block, dense BEV features serve as the input $x$ to the mamba module (Fig.~\ref{fig:occmamba}). By applying discretized SSM dynamics and global convolution kernels, the mamba block effectively processes the BEV features, resulting in an enhanced feature representation with richer long-distance dependencies. This enhanced representation improves the performance of semantic occupancy prediction by capturing and strengthening the spatial relationships between semantic elements. Meanwhile, mamba block's parallel feature learning properties ensure real-time reasoning in dynamic environments.

\noindent\textbf{\textit{Geometry Branch and Geo-Mamba Block}:} The geometric branch (Fig.~\ref{fig:occmamba}) begins with an input layer using a $7 \times 7 \times 7$ kernel and comprises three Geo-Mamba blocks as the encoder. Each Geo-Mamba block maintains a consistent architecture, combining a residual block with a mamba module and a BEV projection module. The residual block first processes the voxels $V \in \mathbb{R}^{1 \times L \times W \times H}$ obtained from the point cloud, and its output features serve as the input $x$ to the mamba module. The mamba module generates multi-scale dense geometric features $\{G_f^1, G_f^2, G_f^3\}$, which enrich the captured geometric details. By leveraging the mamba module's ability to capture long-distance dependencies with linear complexity, the geometric branch can effectively process and refine the geometric information within the voxels. Subsequently, the dense 3D features are aligned along the $z$ axis, and 2D convolutions are applied to compress the feature dimensions. This step produces dense BEV features $\{BEV_{f}^{com,1}, BEV_{f}^{geo,2}, BEV_{f}^{geo,3}\}$, which are suitable for fusion with the semantic features (Fig.~\ref{fig:occmamba}).

\noindent\textbf{\textit{BEV Feature Fusion Branch}:} Our BEV fusion branch adopts a U-Net architecture with 2D convolutions. The encoder consists of an input layer and four residual blocks, where the resolution of the first three residual blocks corresponds to the resolution of the semantic and geometric branches. After each residual block, a feature fusion module from \cite{mei2023ssc} is employed to take the output of the previous stage and the semantic/geometric representation at the same scale as the input. This module outputs fused features that contain informative semantic context and geometric structure. The decoder upscales the compressed features from the encoder three times, through skip connections. These skip connections allow the decoder to recover spatial details that may have been lost during the encoding process, thereby improving the overall quality of the output. Finally, the last convolution layer of the decoder generates the 3D semantic occupancy prediction: $\mathcal{O}\in \mathbb{R}^{((C_n+1)*L)*H*W}$, where $C_n$ is the number of semantic classes.

\subsection{Optimization}

Based above of our methods above, there are four main terms in our loss function. The semantic loss $L_{sem}$ is the sum of the lovasz loss \cite{berman2018lovasz} and cross-entropy loss \cite{zhang2018generalized} at each stage of the semantic branch:
\begin{equation}
 L_{sem}=\sum_{i=1}^{3}(L_{cross,i} + L_{lovasz,i}) 
\end{equation}
The training loss $L_{com}$ for this branch is calculated as follows:
\begin{equation}
  L_{com}=\sum_{i=1}^{3}(L_{binary\_cross,i} + L_{lovasz,i}) 
\end{equation}
where $i$ denotes the $i-th$ stage of the completion branch and $L_{binary\_cross}$ indicates the binary cross-entropy loss. The BEV loss $L_{bev}$ is :
\begin{equation}
 L_{bev}=3\times  (L_{cross} + L_{lovasz})
\end{equation}
We train the whole network end-to-end. The overall objective function is:
\begin{equation}
 L_{total}=L_{bev} +  L_{sem} + L_{com}
\end{equation}
where $L_{bev}$, $ L_{sem}$ and $L_{com}$ respectively represent BEV loss, the semantic loss and completion loss.

\section{AGR Motion Planner}
\label{sec:4}
We introduce AGR-Planner, a novel gradient-based local planner tailored for AGRs built upon EGO-Swarm \cite{zhou2021ego}. It features a Kinodynamic A* algorithm for efficient pathfinding and a gradient-based method for trajectory optimization, streamlining the planning process.

\subsection{Kinodynamic Hybrid A* Path Searching}

Our AGR-Planner begins by generating a preliminary "initial trajectory" $\iota$ (see Fig.~\ref{fig:planner}a), which initially disregards obstacles by incorporating random coordinate points, anchored by the start and end locations. Subsequently, for any "collision trajectory segment" found within obstacles, we employ the kinodynamic A* algorithm to create a "guidance trajectory segment" $\tau$. This segment is defined using motion primitives rather than straight lines for edges during the search, incorporating additional energy consumption metrics for flying (Fig.~\ref{fig:planner}a). This approach encourages the planning of ground trajectories, resorting to aerial navigation only when confronting significant obstacles, thus optimizing energy efficiency.

\subsection{Gradient-Based B-spline Trajectory Optimization}
\begin{figure}[t]
  \centering
     \includegraphics[width=0.85\linewidth]{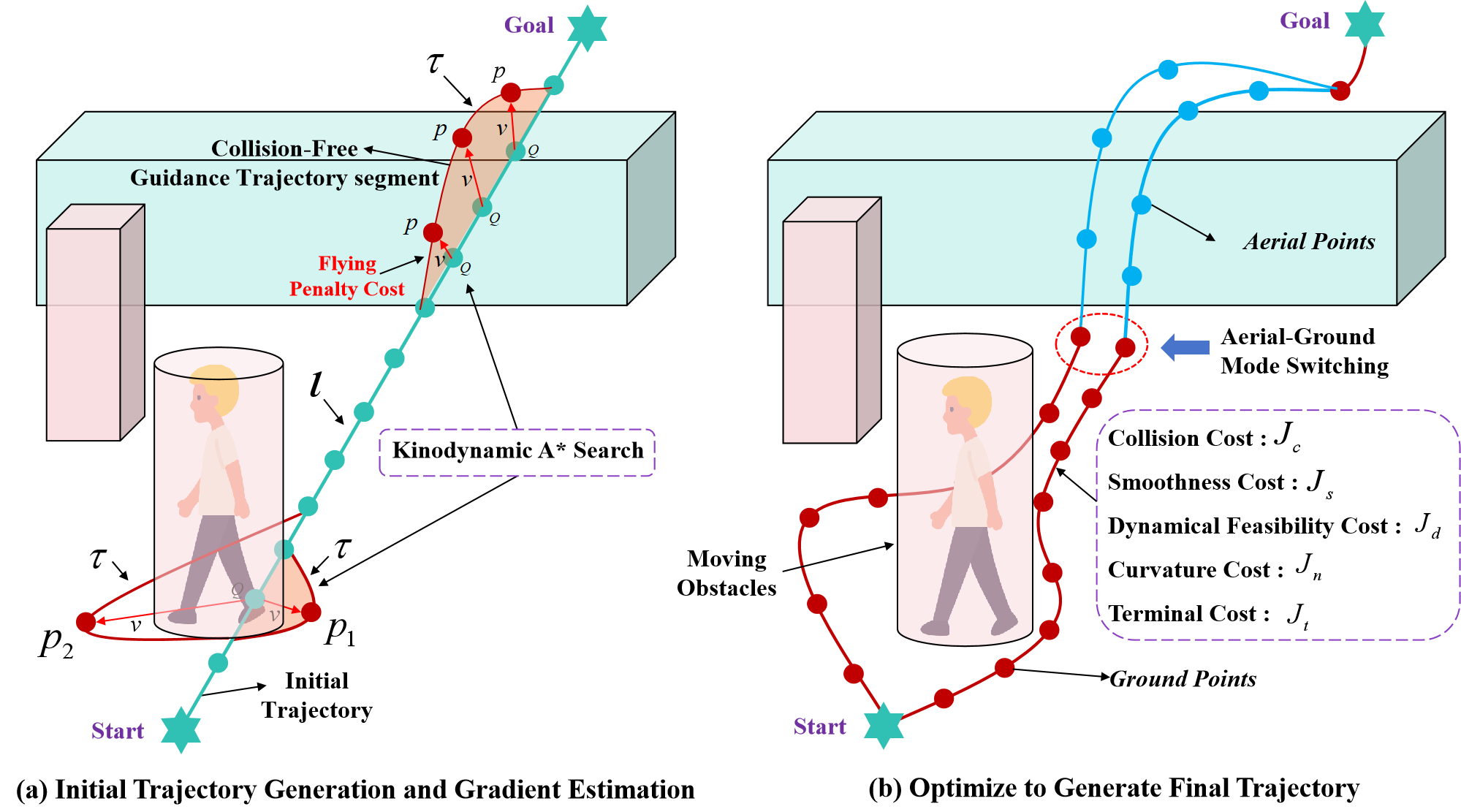}
   \caption{\small AGR-Planner and topological trajectory generation. }
   \label{fig:planner}
\end{figure}

\noindent\textbf{\textit{B-spline Trajectory Formulation}:} In trajectory optimization (Fig.~\ref{fig:planner}b),  the trajectory is parameterized by a uniform B-spline curve $\Theta$, which is uniquely determined by its degree $p_b$, $N_c$ control points $\left \{ Q_1,Q_2,Q_3,...,Q_{N_c} \right \} $, and a knot vector $\left \{ t_1,t_2,t_3,...,t_{M-1},t_M \right \} $, where $Q_i\in \mathbb{R}^3,t_m\in\mathbb{R},M=N+p_b$. Following the matrix representation of the \cite{wang2024agrnav} the value of a B-spline can be evaluated as:
\begin{equation}
 \Theta(u)=[1, u,...,u^p]\cdot M_{p_b+1} \cdot[Q_{i-p_b},Q_{i-p_b+1},...,Q_i]^T 
\end{equation}
where $M_{p_{b}+1}$ is a constant matrix depends only on $p_b$. And $\small u=(t-t_i)/(t_{i+1}-t_i)$, for $t\in [t_i,t_{i+1})$. In particular, in ground mode, we assume that AGR is driving on flat ground so that the vertical motion can be omitted and we only need to consider the control points in the two-dimensional horizontal plane, denoted as $Q_{ground}=\left \{ Q_{t0},Q_{t1},...,Q_{tM} \right \} $, where $Q_{ti}=(x_{ti},y_{ti}),i\in \left [ 0,M \right ]$. In aerial mode, the control points are denoted as $Q_{aerial}$. According to the properties of B-spline: the $k^{th}$ derivative of a B-spline is still a B-spline with order $p_{b,k}=p_b-k$, since $\bigtriangleup t$  is identical alone $\Theta$, the control points of the velocity $V_i$, acceleration $A_i$ and jerk $J_i$ curves are obtained by:
\begin{equation}
 V_i=\frac{Q_{i+1}-Q_i}{\bigtriangleup t},A_i=\frac{V_{i+1}-V_i}{\bigtriangleup t}, J_i=\frac{A_{i+1}-A_i}{\bigtriangleup t}  
\end{equation}

\noindent\textbf{\textit{Collision Avoidance Force Estimation}:} For each control point on the collision trajectory segment, vector $v$ (i.e., a safe direction pointing from inside to outside of that obstacle) is generated from $\iota$ to $\tau$ and $p$ is defined at the obstacle surface (in Fig.~\ref{fig:planner}a). With generated $\left \{ p,v \right \} $ pairs, the planner maximizes $D_{ij}$ and returns an optimized trajectory. The obstacle distance $D_{ij}$ if $i^{th}$ control point $Q_i$ to $j^{th}$ obstacle is defined as:
\begin{equation}
D_{ij}=(Q_i-p_{ij})\times v_{ij}
\end{equation}
Because the guide path $\tau$ is energy-saving, the generated path is also energy efficient (in Fig.~\ref{fig:planner}a). Inspired by \cite{zhou2021ego}, we discover multiple viable paths that navigate through different local minima in a dynamic environment. This is achieved by generating distance fields in various directions through reversing vector $v_1$ to obtain $v_{2}=-v_1$. Following this, a search process identifies a new anchor point $p_2$ on the obstacle's surface along $v_{2}$, as shown in Fig.~\ref{fig:planner}a. This approach enables us to evaluate alternative trajectories and select the most cost-effective path for navigation.

\noindent\textbf{\textit{Air-Ground Hybrid Trajectory Optimization}:} Based on the special properties of AGR bimodal, we first adopt the following cost terms designed by \textit{Zhou et al.} \cite{zhou2021ego}:
\begin{equation}
\min_{Q} J_{AGR}=\sum \lambda _\phi J_\phi 
\end{equation}
where $\phi =\left \{ s,c,d,t \right \} $ and the subscripted  $\lambda$ indicates the corresponding weights. In addition, $J_s$ is the smoothness penalty, $J_c$ is for collision, $J_d$ is for dynamically
feasibility, and $J_t$ is for terminal progress. $\lambda_s,\lambda_c,\lambda_d,\lambda_t$ are weights for each cost terms. Meanwhile, $J_s$ and $J_t$ belongs to \textit{minimum error} which minimize the total error between a linear transformation of decision variables $L(Q)$ and a desired value $D$. $J_c$ and $J_d$ belongs to \textit{soft barrier constraint}  which penalize decision variables exceeding a specific threshold $\zeta $. Subsequently, based on our observations, AGR also faces non-holonomic constraints when driving on the ground, which means that the ground velocity vector of AGR must be aligned with its yaw angle. Additionally, AGR needs to deal with curvature limitations that arise due to minimizing tracking errors during sharp turns. Therefore, a cost for curvature needs to be added, and $J_n$ can be formulated as:
\begin{equation}
 J_n=\sum_{i=1}^{M-1}F_n(Q_{ti}) 
\end{equation}
where $F_n(Q_{ti})$  is a differentiable cost function with $C_{max}$ specifying the curvature threshold:
\begin{equation}
F_n(Q_{ti})=\left\{\begin{matrix} 
  (C_i-C_{max})^2,C_i>C_{max}, \\  
  0,C_i\le C_{max} 
\end{matrix}\right. 
\end{equation}
where $C_i=\frac{\bigtriangleup \beta _i}{\bigtriangleup Q_{ti}} $ is the curvature at $Q_{ti}$, and the $\small \bigtriangleup \beta _i=\left | \tan ^{-1} \frac{\bigtriangleup y_{ti+1}}{\bigtriangleup x_{ti+1}} - \tan ^{-1} \frac{\bigtriangleup y_{ti}}{\bigtriangleup x_{ti}}\right | $ . In general, the overall objective function is formulated as follows:
\begin{equation}
\min_{Q} J_{AGR} = \lambda_sJ_s + \lambda_cJ_c + \lambda_dJ_d  + \lambda_tJ_t + \lambda_n J_n\\
\end{equation}

The optimization problem is addressed with the NLopt. Meanwhile, the same methods as in \cite{wang2024agrnav} are used for trajectory tracking and control, as well as additional mode switching.

\section{Evaluation}
\label{sec:5}
We evaluate OccMamba on the SemanticKITTI benchmark and integrate the best pre-trained model with AGR-Planner to establish our comprehensive \textbf{\name} system. We then assess \name's autonomous navigation efficiency for AGRs in simulated and real-world dynamic environments, focusing on metrics like planning success rate, average movement and planning time, and energy consumption. Finally, ablation experiments verify the navigation efficiency improvements brought by OMEGA's two key components.

\begin{table*}[t]
		\footnotesize
		\setlength{\tabcolsep}{0.004\linewidth}
		\caption{3D Semantic occupancy prediction results on SemanticKITTI test set. The C and L denote Camera and LiDAR.}
		
		\newcommand{\classfreq}[1]{{~\tiny(\semkitfreq{#1}\%)}}  %
		\centering
		\begin{tabular}{l|c|c c| c c c c c c c c c c c c c c c c c c c | c}
			\toprule
			Method
                & \makecell[c]{Modality}
			& IoU $\uparrow$ & mIoU $\uparrow$
			& \rotatebox{90}{\textcolor{road}{$\blacksquare$} road}
			\rotatebox{90}{\ \ \ \classfreq{road}} 
			& \rotatebox{90}{\textcolor{sidewalk}{$\blacksquare$} sidewalk}
			\rotatebox{90}{\ \ \ \classfreq{sidewalk}}
			& \rotatebox{90}{\textcolor{parking}{$\blacksquare$} parking}
			\rotatebox{90}{\ \ \ \classfreq{parking}} 
			& \rotatebox{90}{\textcolor{other-ground}{$\blacksquare$} other-grnd}
			\rotatebox{90}{\ \ \ \classfreq{otherground}} 
			& \rotatebox{90}{\textcolor{building}{$\blacksquare$}  building}
			\rotatebox{90}{\ \ \ \classfreq{building}} 
			& \rotatebox{90}{\textcolor{car}{$\blacksquare$}  car}
			\rotatebox{90}{\ \ \ \classfreq{car}} 
			& \rotatebox{90}{\textcolor{truck}{$\blacksquare$}  truck}
			\rotatebox{90}{\ \ \ \classfreq{truck}} 
			& \rotatebox{90}{\textcolor{bicycle}{$\blacksquare$}  bicycle}
			\rotatebox{90}{\ \ \ \classfreq{bicycle}} 
			& \rotatebox{90}{\textcolor{motorcycle}{$\blacksquare$} motorcycle}
			\rotatebox{90}{\ \ \ \classfreq{motorcycle}} 
			& \rotatebox{90}{\textcolor{other-vehicle}{$\blacksquare$}  other-veh.}
			\rotatebox{90}{\ \ \  \classfreq{othervehicle}} 
			& \rotatebox{90}{\textcolor{vegetation}{$\blacksquare$} vegetation}
			\rotatebox{90}{\ \ \ \classfreq{vegetation}} 
			& \rotatebox{90}{\textcolor{trunk}{$\blacksquare$}  trunk}
			\rotatebox{90}{\ \ \ \classfreq{trunk}} 
			& \rotatebox{90}{\textcolor{terrain}{$\blacksquare$} terrain}
			\rotatebox{90}{\ \ \ \classfreq{terrain}} 
			& \rotatebox{90}{\textcolor{person}{$\blacksquare$}  person}
			\rotatebox{90}{\ \ \ \classfreq{person}} 
			& \rotatebox{90}{\textcolor{bicyclist}{$\blacksquare$}  bicyclist}
			\rotatebox{90}{\ \ \ \classfreq{bicyclist}} 
			& \rotatebox{90}{\textcolor{motorcyclist}{$\blacksquare$}  motorcyclist.}
			\rotatebox{90}{\ \ \ \classfreq{motorcyclist}} 
			& \rotatebox{90}{\textcolor{fence}{$\blacksquare$} fence}
			\rotatebox{90}{\ \ \ \classfreq{fence}} 
			& \rotatebox{90}{\textcolor{pole}{$\blacksquare$} pole}
			\rotatebox{90}{\ \ \ \classfreq{pole}} 
			& \rotatebox{90}{\textcolor{traffic-sign}{$\blacksquare$} traf.-sign}
			\rotatebox{90}{\ \ \ \classfreq{trafficsign}} 
                & FPS
			\\
			\midrule

    	MonoScene~\cite{cao2022monoscene}&C &34.2 & 11.1 & 54.7 & 27.1 &24.8 & 5.7 & 14.4 & 18.8 & 3.3 & 0.5 & 0.7 & 4.4& 14.9 & 2.4 & 19.5 & 1.0 & 1.4 & 0.4 & 11.1 & 3.3 & 2.1 & 1.1 \\
			
            
            OccFormer~\cite{zhang2023occformer} & C &34.5  &12.3& 55.9& 30.3& 31.5& 6.5 &15.7 &21.6& 1.2& 1.5& 1.7 &3.2& 16.8& 3.9 &21.3& 2.2& 1.1& 0.2 &11.9 &3.8 &3.7  &1.8  \\
            
            VoxFormer~\cite{li2023voxformer} & C & 43.2 & 13.4 & 54.1 & 26.9 & 25.1 & 7.3 & 23.5 & 21.7 &3.6 & 1.9 & 1.6 & 4.1 & 24.4 & 8.1 & 24.2 & 1.6 & 1.1 & 0.0 & 6.6 & 5.7 & 8.1 &1.5\\

            TPVFormer~\cite{huang2023tri} & C &34.3 & 11.3 & 55.1 & 27.2 & 27.4 & 6.5 & 14.8 & 19.2 & 3.7 & 1.0 & 0.5 & 2.3 & 13.9 & 2.6 & 20.4 & 1.1 & 2.4 & 0.3 & 11.0 & 2.9 & 1.5 & 1.0\\
             
            LMSCNet~\cite{roldao2020lmscnet} &L &55.3 &17.0&64.0& 33.1 &24.9& 3.2& 38.7& 29.5 &2.5& 0.0& 0.0& 0.1& 40.5& 19.0 &30.8& 0.0& 0.0& 0.0& 20.5& 15.7& 0.5 & 21.3 \\
            
            SSC-RS~\cite{mei2023ssc}&L &59.7   & 24.2& \textbf{73.1} & 44.4 & 38.6 &17.4 &\textbf{44.6} &36.4 &5.3 &10.1 &5.1& \textbf{11.2}& 44.1 &26.0 &41.9 &4.7& 2.4& 0.9& 30.8& 15.0& 7.2 & 16.7\\

            SCONet~\cite{wang2024agrnav}&L &56.1   & 17.6& 51.9 & 30.7 & 23.1 & 0.9 & 39.9 &29.1 &1.7 &0.8 &0.5& 4.8& 41.4 &27.5 &28.6 &0.8& 0.5& 0.1& 18.9& 21.4& 8.0 & 20.0\\
            
            
            M-CONet~\cite{wang2023openoccupancy}&C\&L &55.7 & 20.4 & 60.6 & 36.1 &29.0  & 13.0 & 38.4 &33.8 & 4.7 &3.0 &2.2  & 5.9 & 41.5 &20.5 &35.1 & 0.8 & 2.3 & \textbf{0.6} & 26.0 &18.7 & 15.7  &1.4 \\
                   
            Co-Occ~\cite{pan2024co} &C\&L &56.6 &24.4  & 72.0  &43.5 & 42.5 & 10.2 &35.1  & \textbf{40.0}& \textbf{6.4} &4.4 &3.3 &8.8&41.2&\textbf{30.8}& 40.8 & 1.6 & \textbf{3.3} & 0.4 & \textbf{32.7}& \textbf{26.6}& \textbf{20.7}  & 1.1  \\
	      \midrule

             OccMamba (Ours)&L &\textbf{59.9} &\textbf{25.0}  & 72.9  & \textbf{44.8} &\textbf{42.7}  & \textbf{18.1} & 44.2  & 36.1 & 3.5 & \textbf{12.3} & \textbf{6.0}  &10.1  & \textbf{44.6}  & 29.5  &\textbf{42.1} & \textbf{5.9} & 2.9 & 0.4 & 32.2 & 17.6 & 8.1 & \textbf{22.1}  \\
   
   \bottomrule
		\end{tabular}
		
		\vspace{1mm}
		
		\label{tab:kitti}
		
	\end{table*}

\begin{table}[t]
\centering
\caption{3D occupancy results on SemanticKITTI \cite{behley2019semantickitti} validation set.}
\resizebox{\columnwidth}{!}{%
\begin{tabular}{@{}lccccccc@{}}
\toprule
Method & IoU (\%) & mIoU (\%) & Prec. (\%) & Recall (\%) & Params(M) & FLOPs(G) & Mem. (GB) \\ 
\midrule
\multicolumn{8}{c}{\emph{MLP/CNN-based}} \\
\midrule
Monoscene \cite{cao2022monoscene} & 37.1 & 11.5 & 52.2 & 55.5 & 149.6 & 501.8 & 20.3 \\
NDC-Scene \cite{yao2023ndc}  & 37.2 & 12.7 & - & - & - & - & 20.1 \\
Symphonies \cite{jiang2023symphonize}  & 41.9 & 14.9 & 62.7 & 55.7 & 59.3 & 611.9 & 20.0 \\
\midrule
\multicolumn{8}{c}{\emph{Transformer-based}} \\
\midrule
OccFormer \cite{zhang2023occformer} & 36.5 & 13.5  & 47.3  & 60.4    & 81.4  & 889.0  & 21.0\\
VoxFormer \cite{li2023voxformer} &57.7 & 18.4  & 69.9  & 76.7   & 57.8  & - & 15.2  \\
TPVFormer \cite{huang2023tri}& 35.6 & 11.4  & - & -  & 48.8  & 946.0  & 20.0 \\
CGFormer \cite{yu2024context}& 45.9 & 16.9  &62.8  &63.2   & 122.4  & \textbf{314.5}  & 19.3 \\
\midrule
\multicolumn{8}{c}{\emph{Mamba-based (Ours)}} \\
\midrule
\textbf{OccMamba} & \textbf{58.6} & \textbf{25.2} & \textbf{77.8} & 70.5 & \textbf{23.8} & 505.1 & \textbf{3.5} \\
\bottomrule
\end{tabular}%
}
\label{tab:3d_occupancy_results}
\end{table}

\begin{figure}[htp]
  \centering
     \includegraphics[width=0.8\linewidth]{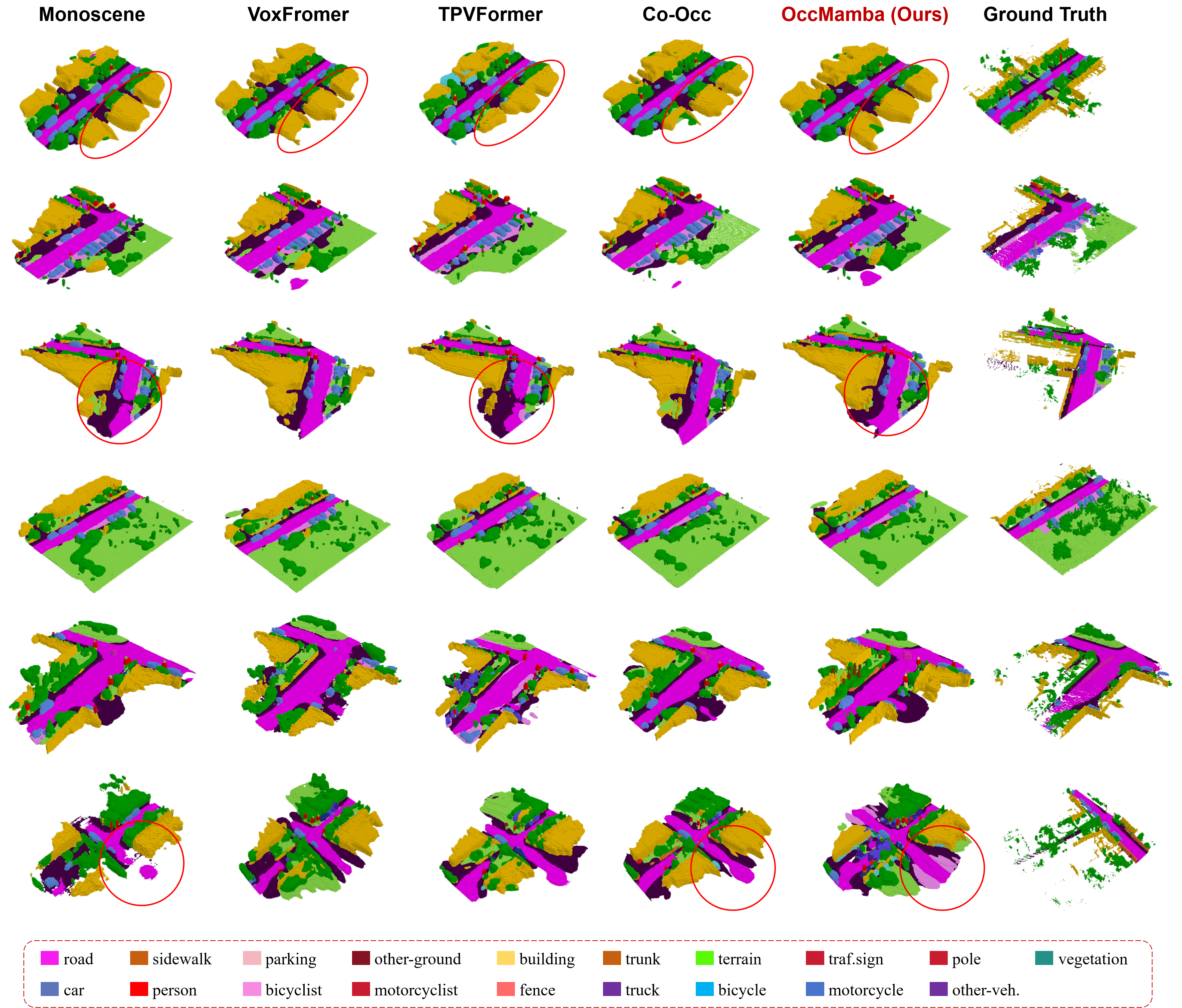}
   \caption{\small Results of a qualitative comparison on the SemanticKITTI validation set are presented, showcasing various models.}
   \label{fig:ssc}
\end{figure}

\subsection{Evaluation setup} 
\noindent\textbf{\textit{3D Semantic Occupancy Network:}} We trained OccMamba on the outdoor SemanticKITTI dataset \cite{behley2019semantickitti} using a single NVIDIA 3090 GPU. The dataset provides LiDAR point clouds for semantic occupancy prediction, with ground truth represented as $[256, 256, 32]$ voxel grids (each voxel measuring $0.2m^3$). Training spanned 80 epochs with a batch size of 6, utilizing the AdamW optimizer \cite{kingma2014adam} with an initial learning rate of 0.001. We applied random x-y axis flipping for input point cloud augmentation. The total training time was approximately 40 hours.

\begin{figure}[t]
  \centering
  \includegraphics[width=0.9\linewidth]{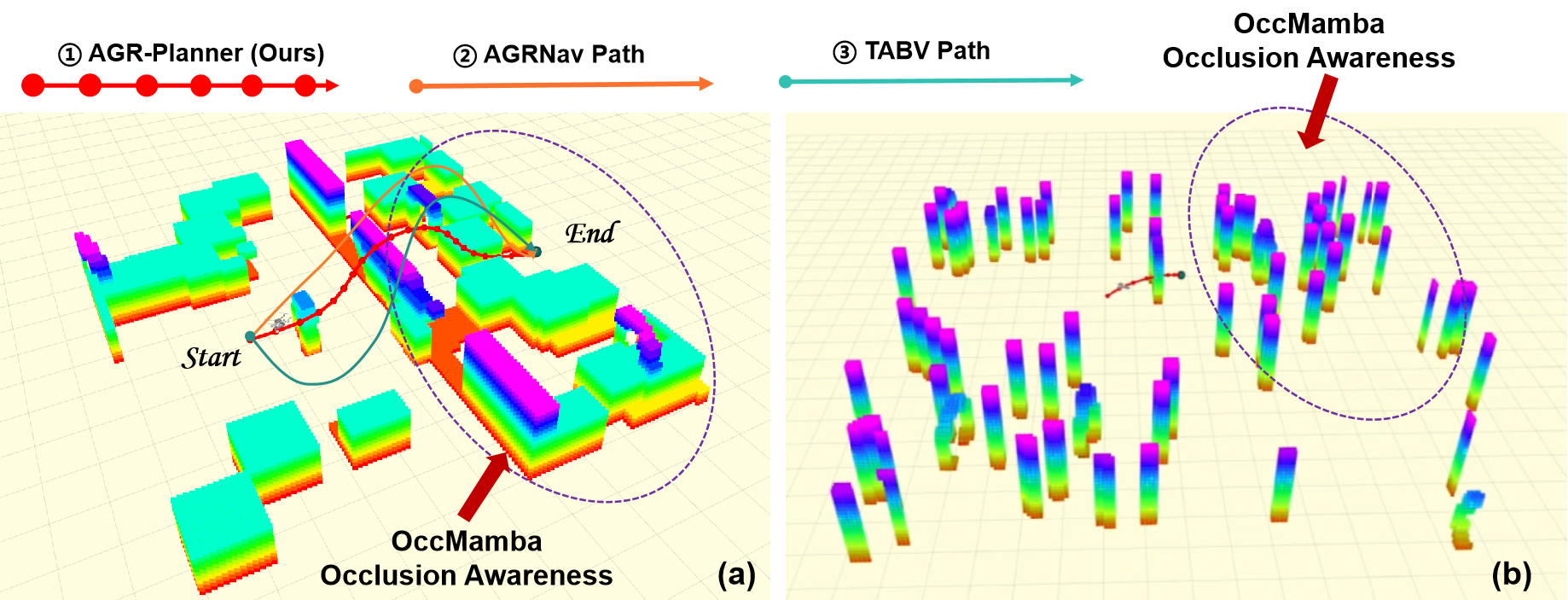}
  \vspace{0.2em} 
  \includegraphics[width=\linewidth]{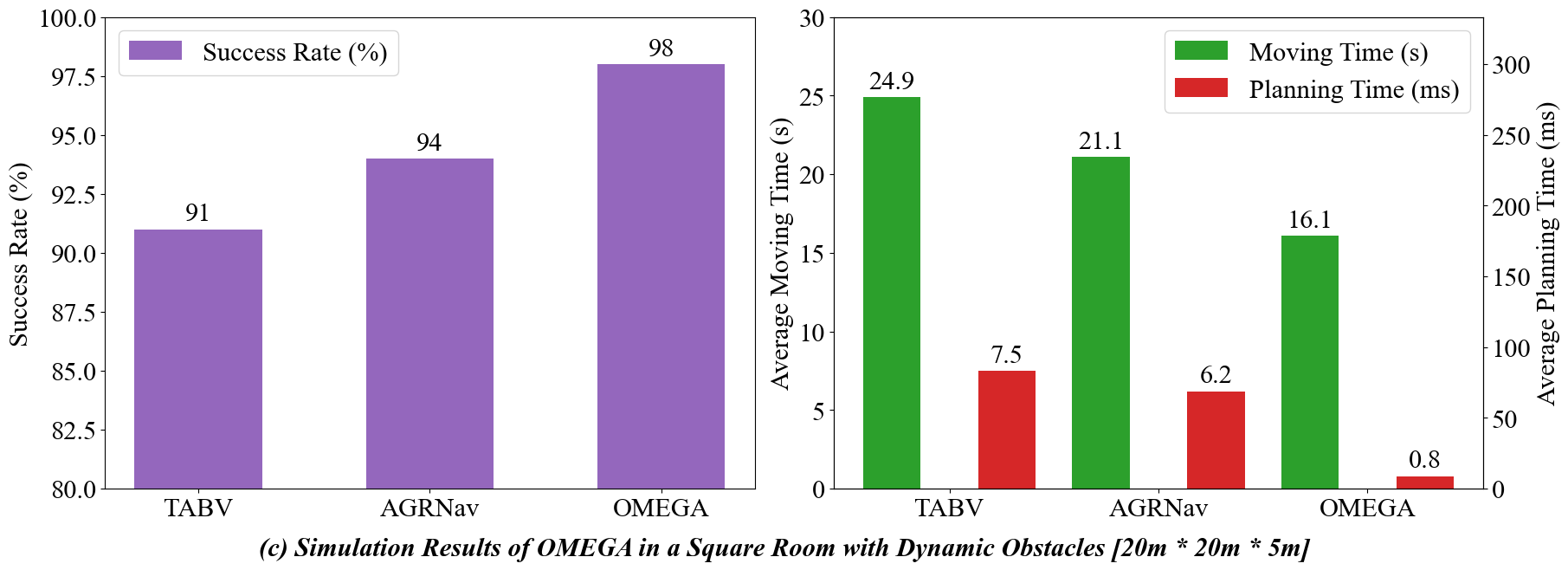}
  \caption{\small Comparative Analysis of \name with Baseline Systems AGRNav \cite{wang2024agrnav} and TABV \cite{zhang2022autonomous}: Quantitative and Qualitative Insights in Simulation Environments.}
  \label{fig:sim_combined}
\end{figure}

\noindent\textbf{\textit{Simulation Experiment:}} We conducted our simulation experiments on a laptop running Ubuntu 20.04 with an NVIDIA RTX 4060 GPU. The experiments took place in a $20m \times 20m \times 5m$ simulated environment, and a total of 200 trials were performed. The first set of 100 trials featured an environment populated with 80 walls and 20 rings, while the second set of 100 trials included 80 moving cylinders and 20 stationary rings. To generate a diverse range of occlusions and unknown areas, the positions of the obstacles were varied in each trial (Fig.~\ref{fig:sim_combined}). The objective was for the AGR equipped with OMEGA to navigate through these environments while avoiding collisions.

\noindent\textbf{\textit{Real-world Experiment:}} In our real-world experiments, \name was deployed on a custom AGR platform (Fig.~\ref{fig:overview}) equipped with a RealSense D435i for point cloud acquisition, a T265 camera for visual-inertial odometry, and a  for real-time onboard computation. The best-performing pre-trained OccMamba model (IoU=59.9) was deployed offline, using the same settings from  AGRNav \cite{wang2024agrnav} . Optimized with TensorRT on Jetson Xavier NX, it efficiently completed local maps for smooth navigation.  





\subsection{OccMamba Comparison against the state-of-the-art.}
\label{sec:B1}
\noindent\textbf{\textit{Quantitative Results:}} OccMamba has achieved state-of-the-art performance on the SemanticKITTI hidden test dataset, boasting a completion IoU of 59.9\% and a mIoU of 25.0\% (Table~\ref{tab:kitti}). While recent methods (e.g., M-CONet \cite{wang2023openoccupancy} and Co-Occ\cite{pan2024co}) have shown commendable performance in semantic occupancy prediction tasks, OccMamba outperforms the latter by a notable margin of 5.8\% in IoU. This leap in accuracy is achieved solely by utilising point cloud data, without the need for other modalities inputs, simplifying integration and deployment in real-world robotic applications. In addition, OccMamba's efficiency parallels its performance; it integrates a linear-time mamba block within a compact, multi-branch network architecture and employs BEV feature fusion, culminating in a lean framework, weighing only 23.8MB (Table~\ref{tab:3d_occupancy_results}). Meanwhile, OccMamba operates on a modest 3.5GB of GPU memory per training batch and impresses with an inference speed of 22.1 FPS , markedly outpacing existing methods like Monoscene\cite{cao2022monoscene} and TPVFormer\cite{huang2023tri} by a factor of 20.

\noindent\textbf{\textit{Qualitative Results:}} In the visualizations on the SemanticKITTI validation set (Fig.~\ref{fig:occmamba}), OccMamba demonstrates excellent semantic occupancy predictions in occlusion areas, particularly for complex or moving categories such as ``vegetation", ``terrain", ``person", and ``bicycle", aligning with the quantitative results in Table~\ref{tab:kitti}. These reliable predictions are essential for subsequent path planning.

\begin{table}[htp]
\caption{\small Ablation Study on SemanticKITTI  Validation Set.}
\small
\centering
\resizebox{\columnwidth}{!}{%
\begin{tabular}{lccccc}
\toprule
Method & IoU $\uparrow$ & mIoU $\uparrow$ & Prec. & Recall & F1 \\
\midrule
OccMamba & 58.6 & 25.2 & 77.8 & 70.5 & 73.9 \\
w/o Geo-Mamba Block & 58.2 & 24.7 & 76.4 & 69.8 & 72.8 \\
w/o Sem-Mamba Block & 57.8 & 24.1 & 76.1 & 69.5 & 72.5 \\
\bottomrule
\end{tabular}%
}
\label{tab:ablation}
\end{table}

\noindent\textbf{\textit{Ablation Study:}} The ablation experiments on the SemanticKITTI validation set (Table~\ref{tab:ablation}) underscore the significance of the Sem-Mamba and Geo-Mamba blocks in our network architecture. Separating semantic and geometric feature processing promotes focused representation learning and leverages their synergistic effects. Removing the Sem-Mamba block results in a substantial 4.37\% drop in mIoU, highlighting its vital role in accurate semantic segmentation and nuanced scene element classification.

\subsection{Simulated Air-Ground Robot Navigation} 
\label{sec:C1}

We conducted a comparative analysis of our \name system, TABV \cite{zhang2022autonomous} and AGRNav\cite{fan2019autonomous}. Through 200 trials with varied obstacle placements, we evaluated the average moving time, planning time, and success rate (i.e., collision-free) of each system (Fig.~\ref{fig:sim_combined}c). In dynamic environments, our system \name demonstrates an average planning success rate of 98\% (Fig.~\ref{fig:sim_combined}c). This achievement stems not solely from OccMamba's real-time and precise mapping of occluded areas, which facilitates complete local map acquisition for planning, but also from the innovative design of our AGR-Planner, which plans multiple candidate trajectories without the need for an ESDF map. When benchmarked against two ESDF-reliant navigation baselines, our AGR-Planner reduces planning time by a substantial 89.33\% (Fig.~\ref{fig:sim_combined}c). This synergy between OccMamba's detailed environmental perception and AGR-Planner's efficient path computation guarantees swift and reliable navigation for AGRs in complex, dynamically changing, and visually obstructed scenarios.

\noindent\textbf{\textit{Ablation Study:}} Ablation experiments (Table~\ref{tab:ablation-experiments}) demonstrate that OccMamba contributes more to OMEGA's planning success than SCONet, improving it by 4\% and 1\%, respectively. Moreover, replacing the SCONet in \cite{wang2024agrnav} with OccMamba enhances AGRNav's success rate by 2\%. This experiment highlights the synergy between OccMamba and AGR-Planner in achieving efficient navigation.

\begin{table}[t]
\centering
\caption{\small Ablation Study of OMEGA }
\label{tab:ablation-experiments}
\begin{tabular}{@{}cccc@{}}
\toprule
Percep. & Plan. & Succ. Rate (\% )& Plan. Time (s) \\
\midrule
OccMamba & H-Planner \cite{wang2024agrnav}    & 96   &  6.5 \\
SCONet \cite{wang2024agrnav} & AGR-Planner   & 95 & 0.8 \\
\bottomrule
 - & AGR-Planner & 94  & 0.7 \\
 OccMamba & AGR-Planner  & 98   & 0.8  \\
\bottomrule
\end{tabular}
\end{table}

\begin{figure}[t]
  \centering
  \includegraphics[width=0.9\linewidth]{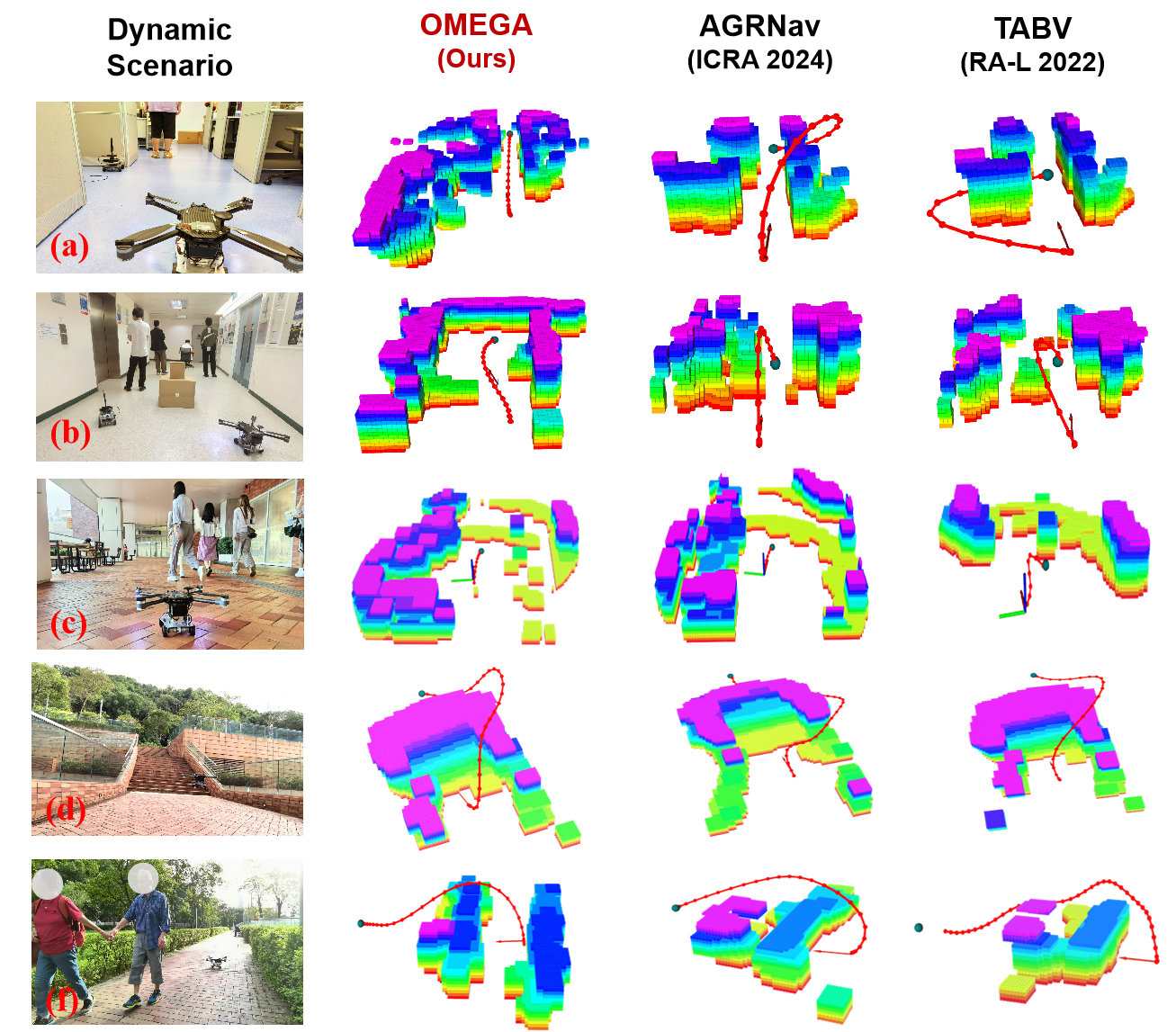}
  \vspace{0.2em} 
  \includegraphics[width=\linewidth]{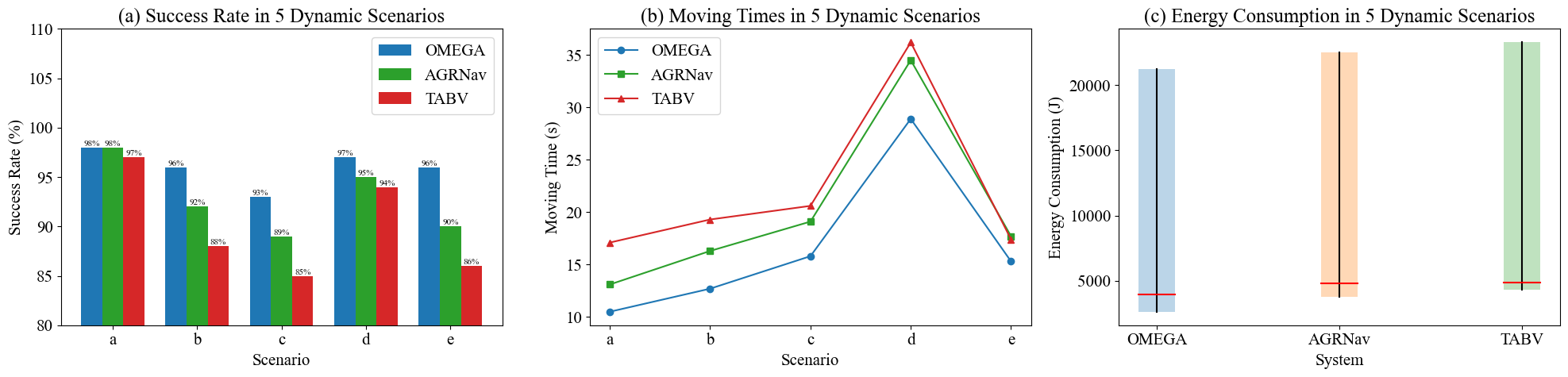}
  \caption{\small  Comparative Analysis of \name with Baseline Systems AGRNav \cite{wang2024agrnav} and TABV \cite{zhang2022autonomous}: Quantitative and Qualitative Insights in 5 real-world dynamic environments. }
  \label{fig:real}
\end{figure}

\subsection{Real-world Air-Ground Robot Navigation}
\label{sec:D1}
We demonstrated OMEGA's superior efficiency and energy conservation in 5 dynamic scenarios with a velocity cap of 1.5 m/s. Each navigation method was tested 10 times per scenario for reliable performance comparison. We achieved a notable 96\% average success rate (Fig.~\ref{fig:real}a) in 5 dynamic scenarios with OMEGA, exhibiting lower average energy consumption relative to competitors (Fig.~\ref{fig:real}c). Specifically, in scenarios B and C, OMEGA recorded energy reductions of 22.1\% and 17.28\%, respectively, against AGRNav. These improvements are largely due to our OccMamba module's swift computational ability to predict obstacle distributions in non-visible areas, enabling the AGR system to avoid potential impediments. When integrated with the AGR-Planner, this foresight allows for the rapid generation of multiple candidate paths, optimizing for both energy efficiency and reduced traversal time (Fig.~\ref{fig:real}b). Please zoom in on Fig.~\ref{fig:real} to see more detailed qualitative and quantitative results.

\vspace{-5pt}
\section{CONCLUSION}
In this letter, we introduce OMEGA, an advanced system for air-ground robot (AGR) navigation in dynamic environments. OMEGA features OccMamba, which efficiently generates comprehensive, occlusion-free local maps using mamba blocks for real-time semantic occupancy mapping. The AGR-Planner component then utilizes these maps to produce safe and dynamically feasible trajectories. Our evaluations show OccMamba outperforming state-of-the-art methods with 59.9\% IoU at 22.1 FPS. When integrated into OMEGA, the system achieves a 96\% average success rate in real-world dynamic scenarios.

\ifCLASSOPTIONcaptionsoff
  \newpage
\fi

{\small
\bibliographystyle{IEEEtran}
\balance
\bibliography{IEEEfull}
}

\end{document}